\documentclass{article}

\usepackage{proceed2e,url}
\usepackage[normalem]{ulem}
\usepackage{multirow}
\usepackage{color}
\usepackage{rotating}

\usepackage{amsmath, amsthm, amssymb}

\newcommand{\x}[1]{\ensuremath{I(#1)}}
\newcommand{\fv}[2]{\ensuremath{\x{#1 \rightarrow #2}}}

\newcommand{\xx}[1]{\ensuremath{J(#1)}}
\newcommand{\fvfv}[2]{\ensuremath{\xx{#1 \rightarrow #2}}}

\newcommand{\localscore}[2]{\ensuremath{c(#1,#2)}}

\newcommand{\varsubset}{\ensuremath{W}}
\newcommand{\cluster}{\ensuremath{C}}

\newcommand{\var}{\ensuremath{v}}

\newcommand{\varrr}{\ensuremath{w}}

\newcommand{\lrs}{\textsf{lrs}}
\newcommand{\gobnilp}{\textsf{GOBNILP}}
\newcommand{\oldgobnilp}{\textsf{GOBNILP 1.0}}
\newcommand{\newgobnilp}{\textsf{GOBNILP 1.3}}

\newcommand{\scip}{\textsf{SCIP}}
\newcommand{\cplex}{\textsf{CPLEX}}
\newcommand{\urlearning}{\textsf{URLearning}}

\newcommand{\lpsol}{\ensuremath{x^*}}
\newcommand{\lpsolhalf}{\ensuremath{y^*}}
\newcommand{\lpsolfour}{\ensuremath{z^*}}
\newcommand{\lp}[2]{\ensuremath{\lpsol(#1 \rightarrow #2)}}

\newcommand{\nvars}{\ensuremath{p}}

\newcommand{\hull}{\ensuremath{{\cal P}}}
\newcommand{\clusterhull}{\ensuremath{{\cal P}_{cluster}}}
\newcommand{\clusterhullscip}{\ensuremath{{\cal P}'_{cluster}}}
\newcommand{\daghull}{\ensuremath{{\cal P}_{dag}}}

\newcommand{\dimdag}{\ensuremath{n}}

\title{Advances in Bayesian Network Learning using Integer Programming}
\author{\textbf{Mark Barlett}\\Dept of Computer Science \&\\York
  Centre for Complex Systems Analysis\\University of
    York, UK\\ \texttt{mark.bartlett@york.ac.uk} \And \textbf{James Cussens}\\ Dept of Computer Science \&\\York
  Centre for Complex Systems Analysis\\University of
    York, UK\\
    \texttt{james.cussens@york.ac.uk}}
\begin{document}

\maketitle

\begin{abstract}
  We consider the problem of learning Bayesian networks (BNs) from
  complete discrete data. This problem of discrete optimisation is
  formulated as an integer program (IP). We describe the various steps
  we have taken to allow efficient solving of this IP. These are (i)
  efficient search for cutting planes, (ii) a fast greedy algorithm to
  find high-scoring (perhaps not optimal) BNs and (iii) tightening
  the linear relaxation of the IP. After relating this BN
  learning problem to set covering and the multidimensional 0-1
  knapsack problem, we present our empirical results. These show
  improvements, sometimes dramatic, over earlier results.
\end{abstract}

\section{Introduction}
\label{sec:intro}

Bayesian networks (BNs) use a directed acyclic graph (DAG) to represent
conditional independence relations between random variables. Each BN
defines a joint probability distribution over joint instantiations of
the random variables. BNs are a very popular and effective
representation for probabilistic knowledge and there is thus
considerable interest in `learning' them from data (and perhaps also
prior knowledge).

In this paper we present our approach to the `search-and-score' method
of BN learning. As a score for any candidate BN we choose the log
marginal likelihood with Dirichlet priors over the parameters. We
choose these Dirichlet priors so that our score is the well-known BDeu
score. Throughout we set the \emph{effective sample size} for the BDeu
score to be 1. The learning problem is a discrete
optimisation problem: find the BN structure (DAG) with the highest BDeu score.

In Section~\ref{sec:encoding} we present our integer programming (IP)
encoding of the problem. Section~\ref{sec:branchandcut} shows how we
use a `branch-and-cut' approach to solve the IP problem.
Section~\ref{sec:separation} describes how we find `cutting
planes'. Section~\ref{sec:sinks} describes our fast greedy algorithm
for finding good but typically sub-optimal
BNs. Section~\ref{sec:tighter} presents our investigation into the
convex hull of DAGs with 3 or 4 nodes, the facets of which turn out to
be useful for learning DAGs with any number of nodes.
Section~\ref{sec:setcover} is more discursive
in nature: pointing out interesting connections to the set covering
and multi-dimensional knapsack problems. The central
contribution of the paper---faster solving for exact BN learning---is
established in Section~\ref{sec:results} where we present our
empirical results. We finish, as is the custom, with conclusions and
pointers to future work.

This paper assumes familiarity with Bayesian networks and basic
knowledge concerning BN learning. See
\cite{koller09:_probab_graph_model} for a comprehensive treatment of
both these topics and much more besides.  The rest of the paper
assumes the reader knows what an integer program is: it is a discrete
optimisation problem where the objective function is a linear function
of the (discrete) problem variables and where the set of feasible solutions is
defined by a finite set of linear inequalities over these
variables. The \emph{linear relaxation} of the IP is obtained by
removing the constraint that the problem variables take only integer values.
For further reading on integer programming, consult e.g.\
\cite{wolsey98:_integ_progr}.  We refer to
the variables in the learned BN structure as `nodes' rather than `variables' to
more clearly distinguish them from IP variables.

\section{Integer program encoding}
\label{sec:encoding}

We encode our BN structure learning problems as an integer program
(IP) so that the log marginal likelihood of any DAG is a linear
function of the IP variables. This requires creating binary `family'
variables $\fv{\varsubset}{\var}$ for each node \var{} and candidate
parent set \varsubset, where $\fv{\varsubset}{\var} = 1$ iff
\varsubset{} is the parent set for \var. This encoding has been used
in previous work
\cite{cussens08:_bayes_max_sat,jaakkola10:_learn_bayes_networ_struc_lp_relax,cussens11:_bayes_networ_learn_cuttin_planes}. The
objective function then becomes
\begin{equation}
  \label{eq:obj}
\sum_{\var, \varsubset}
\localscore{\var}{\varsubset} \fv{\varsubset}{\var}
\end{equation}
 where
$\localscore{\var}{\varsubset}$ is the `local' BDeu score
for \varsubset{} being the parents of \var.
This score is computed from the
data prior to the search for an optimal BN. Evidently the number of
such scores/variables increases exponentially with $\nvars$, the
number of nodes in the BN. However if $\varsubset \subset \varsubset'$
and $\localscore{\var}{\varsubset} > \localscore{\var}{\varsubset'}$
then, in the absence of any user constraints on the DAG, it is obvious that $\varsubset$ cannot be the parent set for
$\var$ in any optimal BN and thus there is no need to create the unnecessary
variable $\fv{\varsubset'}{\var}$. Moreover, using a pruning
technique devised by de Campos and Ji
\cite{campos10:_proper_bayes_diric_scores_learn} it is possible to
prune the search for necessary variables so that local scores for many
unnecessary variables need never be computed.
Nonetheless, unless $\nvars$ is small
(e.g. $\leq 12$), even with this pruning we typically have to make
some restriction on the number of candidate parent sets for any
node. See Section~\ref{sec:results} for more on this.

The \fv{\varsubset}{\var} variables encode DAGs, but in many cases one
is more interested in \emph{Markov equivalence classes} of DAGs
\cite{koller09:_probab_graph_model} which group together DAGs
representing the same conditional independence relations. This has
motivated the \emph{standard imset} and later \emph{characteristic
  imset} representation \cite{hemmecke12:_charac_bayes}. A
characteristic imset $\mathtt{c}$ is a zero-one vector indexed by
subsets \cluster{} of BN nodes such that $\mathtt{c}(\cluster) = 1$
iff there is a node $\var\in\cluster$ such that all nodes in
$\cluster\setminus\{\var\}$ are parents of \var.  Importantly two DAGs
have the same characteristic imset representation iff they are Markov
equivalent. This representation lends itself to an IP approach to BN
structure learning. Existing  \cite{hemmecke12:_charac_bayes} and
ongoing work shows encouraging results. Connecting the characteristic
imset representation to our `family' representation we have that for
any DAG and any subset of nodes \cluster:
\begin{equation}
  \label{eq:imset}
  \mathtt{c}(\cluster) = \sum_{\var \in \cluster}
\sum_{\varsubset:\cluster \setminus \{\var\} \subseteq \varsubset}
\fv{\varsubset}{\var}
\end{equation}

Returning to our own encoding, to ensure that IP solutions correspond to DAGs two families of linear
constraints are used. Let $V$ be the set of BN nodes.
The convexity constraints
\begin{equation}
  \label{eq:convexity}
  \forall \var\in V: \sum_{\varsubset} \fv{\varsubset}{\var} = 1
\end{equation}
simply ensure that any variable has exactly one parent set. The
`cluster' constraints
\begin{equation}
  \label{eq:cluster}
  \forall \cluster \subseteq V: \sum_{\var \in \cluster}
  \sum_{\varsubset : \varsubset \cap \cluster = \emptyset}
  \fv{\varsubset}{\var} \geq 1
\end{equation}
introduced by Jaakkola \emph{et al}
\cite{jaakkola10:_learn_bayes_networ_struc_lp_relax}, ensure that the
graph has no cycles. For each cluster $\cluster$ the associated
constraint declares that at least one $\var\in\cluster$ has no parents
in \cluster. Since there are exponentially many cluster constraints
these are added as `cutting planes' in the course of solving.

\section{Branch-and-cut algorithm}
\label{sec:branchandcut}

Let \dimdag{} be the number of \fv{\varsubset}{\var} variables so that
any instantiation of these variables can be viewed as a point in
$\{0,1\}^{\dimdag} \subset [0,1]^{\dimdag} \subset
\mathbb{R}^\dimdag$. Let \clusterhull{} be the polytope of all points
in $\mathbb{R}^\dimdag$ which satisfy the bounds on the variables,
convexity constraints (\ref{eq:convexity}) and cluster constraints
(\ref{eq:cluster}).  If a point in \clusterhull{} is entirely
integer-valued (i.e.\ is in $\{0,1\}^{\dimdag}$) then it corresponds to
a DAG, but \clusterhull{} also contains infinitely many
non-DAG points.  Due to the \nvars{} convexity equations
(\ref{eq:convexity}) this polytope is of dimension $\dimdag -
\nvars$. Note that all points in \clusterhull{} have an objective
value (\ref{eq:obj}) not just those with integer values.

Jaakkola \emph{et al}
\cite{jaakkola10:_learn_bayes_networ_struc_lp_relax} consider the
problem of finding a point in \clusterhull{} with maximal objective
value. They consider the dual problem and iteratively add dual
variables corresponding to cluster constraints. This process is
continued until the problem is solved or there are too many dual
variables. In the latter case branch-and-bound is used to continue the
solving process. A `decoding' approach is used to extract DAGs from
values of the dual variables.

We instead take a fairly standard `branch-and-cut' approach to the
problem. The essentials of branch-and-cut are as follows, where
\emph{LP solution} is short for `solution of the current linear
relaxation':

\begin{verbatim}
1. Let x* be the LP solution.
2. If there are valid inequalities
        not satisfied by x*
        add them and go to 1.
   Else if x* is integer-valued then
        the current problem is solved
   Else branch on a variable with
        non-integer value in x*
        to create two new sub-IPs.
\end{verbatim}

The added valid inequalities are called `cutting planes' since they
`cut off' the LP solution \lpsol{}. This process of cutting and
perhaps branching is performed on all nodes in the search tree. If the
objective value of the LP solution of a sub-IP is worse than the current best
integer solution then the tree is pruned at that point. Note that the
term `branch-and-cut' is a little misleading since there is typically
much cutting in the root node before any branching is done.

Where Jaakkola \emph{et al} added dual variables to the dual problem
we add cluster constraints as cutting planes (\emph{cluster cuts}) to
the original `primal' problem. We do a complete search for cluster
cuts so that if none can be found we have a guarantee that the LP
solution \lpsol{} satisfies all cluster constraints
(\ref{eq:cluster}). Note that there is no need to find all cluster
cuts to have this guarantee; in practice only a small fraction of the
exponentially many cluster cuts need be found.

If no cluster cuts can be found (and the problem is not solved) we
search for three other sorts of cutting planes: Gomory, strong
Chv\'{a}tal-Gomory (CG) and zero-half cuts. If none of these cuts can be found we
branch on a fractional $\fv{\varsubset}{\var}$ variable to create two
new sub-IPs as described above.
 Solving the problem returns a guaranteed optimal BN.  Since we
are working with the primal representation there is no decoding
required to extract the BN from the optimal values of the
$\fv{\varsubset}{\var}$ variables: we just return the \nvars{}
variables which are set to 1.

The algorithm is implemented using the the \scip{}
\cite{achterberg07:_const_integ_progr} callable library
(\url{scip.zib.de}). Implementing a basic branch-and-cut algorithm
with \scip{} amounts to little more than setting \scip{} parameter flags
appropriately. We used \scip's built-in functions to search for
Gomory, strong CG and zero-half cuts. We also used
\scip's default approach to branching. See \cite{achterberg07:_const_integ_progr}
and the \scip{} documentation for details. However the search for cluster
cuts we implemented ourselves, plugging it into the branch-and-cut
algorithm as a \scip{} \emph{constraint handler}. This search is
described in Section~\ref{sec:separation}. Our
greedy algorithm for finding good
BNs is implemented as a \scip{} \emph{primal heuristic} and is described in Section~\ref{sec:sinks}.

\section{Finding cluster cuts}
\label{sec:separation}

Our system \newgobnilp{} uses a sub-IP to search for cluster cuts. The approach
is essentially the same as that presented by Cussens
\cite{cussens11:_bayes_networ_learn_cuttin_planes}. Nonetheless we
describe it here for completeness and because our current
implementation has a simple but very effective optimisation that
was missing from the earlier one.

To understand the sub-IP first note that, due to the convexity
constraints, the constraint (\ref{eq:cluster}) for cluster \cluster{} can be reformulated
as a knapsack constraint (\ref{eq:cluster2}):
\begin{equation}
  \label{eq:cluster2}
  \sum_{\var \in \cluster}
  \sum_{\varsubset : \varsubset \cap \cluster \neq \emptyset}
  \fv{\varsubset}{\var} \leq |\cluster| - 1
\end{equation}

In the sub-IP the cluster is represented by $|V|=\nvars$ binary variables
$I(\var\in \cluster)$ each with objective coefficient of -1. Now let
$\lp{\varsubset}{\var}$ be the value of
$\fv{\varsubset}{\var}$ in the LP solution. For each
$\lp{\varsubset}{\var} > 0$ create a sub-IP binary variable
$\fvfv{\varsubset}{\var}$ with an objective coefficient of
$\lp{\varsubset}{\var}$. Abbreviating $\sum_{\var\in V}
I(\var\in \cluster)$ to $|C|$ the sub-IP is defined to be:
\begin{equation}
  \label{eq:opt}
  \mbox{Maximise: } -|C| + \sum \lp{\varsubset}{\var} \fvfv{\varsubset}{\var}
\end{equation}
Subject to, for each \fvfv{\varsubset}{\var}:
\begin{equation}
  \label{eq:c1}
  \fvfv{\varsubset}{\var} \Rightarrow I(\var\in C)
\end{equation}
\begin{equation}
  \label{eq:c2}
  \fvfv{\varsubset}{\var} \Rightarrow \bigvee_{\varrr \in \varsubset} I(\varrr\in C)
\end{equation}
The constraints (\ref{eq:c1}) and (\ref{eq:c2}) are posted as \scip{}
\texttt{logicor} clausal constraints and the sub-IP search is set to
depth-first, branching only on $I(\var\in \cluster)$ variables. Note that \texttt{logicor} constraints are a special type
of linear constraint.
It is a requirement that the
objective value is greater than -1. For efficiency this is implemented
directly as a limit on the objective value (using \scip's
\texttt{SCIPsetObjlimit} function) rather than as a linear
constraint. A final constraint dictates that $|\cluster| \geq 2$.

In any feasible solution we have that
\begin{equation}
  \label{eq:cluster_viol}
  -|\cluster| +
\sum \lp{\varsubset}{\var} \fvfv{\varsubset}{\var} > -1
\end{equation}
Due to the constraints (\ref{eq:c1}) and (\ref{eq:c2}), for
$\fvfv{\varsubset}{\var}$ to be non-zero $\var$ must be in the cluster
\cluster{} and at least one element of \varsubset{} must also be in
\cluster. So (\ref{eq:cluster_viol}) can be rewritten as:
\begin{equation}
  \label{eq:cluster_viol2}
  -|\cluster| +
\sum_{\var \in \cluster}
  \sum_{\varsubset : \varsubset \cap \cluster \neq \emptyset} \lp{\varsubset}{\var} \fvfv{\varsubset}{\var} > -1
\end{equation}
It follows that the cluster \cluster{} associated with a feasible
solution of the sub-IP has a cluster constraint which is violated by
the current LP solution \lpsol. Each feasible solution of the IP thus
corresponds to a valid cutting plane. The sub-IP is always solved to
optimality, (collecting any sub-optimal feasible solutions along the
way) so it follows that if the current LP solution violates any
cluster constraint then at least one will be found.

In
\cite{cussens11:_bayes_networ_learn_cuttin_planes} a sub-IP was also
used to find cutting planes. However, there ``a
binary variable $\fvfv{\varsubset}{\var}$ is created for \textbf{each} family
variable $\fv{\varsubset}{\var}$ in the main IP.'' (our emphasis). In the current approach
$\fvfv{\varsubset}{\var}$ variables are created \emph{only for
  $\fv{\varsubset}{\var}$ variables which are not zero in the LP
  solution}.  This greatly reduces the number of variables in the
sub-IP (and thus speeds up sub-IP solving) since typically most main
IP variables \emph{are} set to zero in any LP solution.

\section{Sink finding algorithm}
\label{sec:sinks}

Finding a good primal solution (i.e.\ a BN) early on in the search is
worthwhile even if it turns out to be suboptimal. High scoring
solutions allow more effective branch-and-bound and may help in the root
node due to \emph{root reduced cost strengthening} \cite[\S
7.7]{achterberg07:_const_integ_progr}. If a problem cannot be solved
to optimality having a good, albeit probably suboptimal, solution is
even more important.

We have a `sink-finding' algorithm which proposes primal solutions
using the current LP solution. The algorithm is based on two ideas:
(i) that there might be good primal solutions `near' the current LP
solution and (ii) that an optimal BN is easily found if we can
correctly guess an optimal total ordering of BN nodes. The first idea
is common to all \emph{rounding heuristics}. \scip{} has 6 built-in
rounding heuristics and we allow \scip{} to run those that are fast
and they do sometimes find high scoring BNs. The second idea has been exploited in dynamic
programming approaches to exact BN learning
\cite{DBLP:conf/uai/SilanderM06}.

To understand how the algorithm works consider
Table~\ref{tab:initial_sink}. The table has a row for each node and
there are $\nvars = |V|$ rows. The $\fv{\varsubset}{\var}$ variables
for each node are ordered according to their objective coefficient
$\localscore{\var}{\varsubset}$, so that, for example,
$\varsubset_{1,1}$ is the `best' parent set for node 1 and
$\varsubset_{1,k_{1}}$ the worst. The objective coefficients
$\localscore{\var}{\varsubset}$ play no role in the sink-finding algorithm
other than to determine this ordering.  Since the number of available
parent sets may differ between nodes the rows will typically not be of
the same length.

\begin{table}
  \centering
\begin{tabular}{|l|l|l|l|}
\hline
\fv{\varsubset_{1,1}}{1} &   \fv{\varsubset_{1,2}}{1} & \dots &
\fv{\varsubset_{1,k_{1}}}{1} \\
\hline
\fv{\varsubset_{2,1}}{2} &   \fv{\varsubset_{2,2}}{2} & \dots &
\fv{\varsubset_{2,k_{2}}}{2} \\
\hline
\fv{\varsubset_{3,1}}{3} &   \fv{\varsubset_{3,2}}{3} & \dots &
\fv{\varsubset_{3,k_{3}}}{3} \\
\hline\dots&\dots & \dots & \dots\\
\hline\fv{\varsubset_{\nvars,1}}{\nvars} &   \fv{\varsubset_{\nvars,2}}{\nvars} & \dots &
\fv{\varsubset_{\nvars,k_{\nvars}}}{\nvars} \\
\hline
\end{tabular}
  \caption{Example initial state of the sink-finding heuristic for
    $|V| = \nvars$. Rows need not be of the same length.}
  \label{tab:initial_sink}
\end{table}

On the first iteration of the sink finding algorithm, for each child
variable the `cost' of selecting its best-scoring parent set is
computed. This cost is $ 1-\lp{\varsubset_{\var,1}}{\var}$,
where $\lp{\varsubset_{\var,1}}{\var}$ is the value of
$\fv{\varsubset_{\var,1}}{\var}$ in the LP solution.

Denote the child variable chosen as $\var_\nvars$. In order to ensure
that the algorithm generates an acyclic graph a total order is also
generated. This is achieved by setting $\fv{\varsubset}{\var}$ to 0 if
$\var_{\nvars} \in \varsubset$. As a result $\var_\nvars$ will be a
sink node of the BN which the algorithm will eventually construct.

\begin{table}
  \centering
\begin{tabular}{|l|l|l|l|}
\hline
\sout{\fv{\varsubset_{1,1}}{1}} &   \fv{\varsubset_{1,2}}{1} & \dots &
\fv{\varsubset_{1,k_{1}}}{1} \\
\hline
$\mathbf{\fv{\varsubset_{2,1}}{2}}$ &   \sout{\fv{\varsubset_{2,2}}{2}} & \sout{\dots} &
\sout{\fv{\varsubset_{2,k_{2}}}{2}} \\
\hline
\fv{\varsubset_{3,1}}{3} &   \sout{\fv{\varsubset_{3,2}}{3}} & \dots &
\fv{\varsubset_{3,k_{3}}}{3} \\
\hline\dots&\dots & \dots & \dots\\
\hline\sout{\fv{\varsubset_{\nvars,1}}{\nvars}} &   \sout{\fv{\varsubset_{\nvars,2}}{\nvars}} & \dots &
\fv{\varsubset_{\nvars,k_{\nvars}}}{\nvars} \\
\hline
\end{tabular}
  \caption{Example intermediate state of the sink-finding heuristic.}
  \label{tab:second_sink}
\end{table}

Suppose that it turned out that $\var_{p}=2$ and that node 2 was a member
of parent sets $\varsubset_{1,1}$, $\varsubset_{3,2}$, $\varsubset_{\nvars,1}$ and
$\varsubset_{\nvars,2}$. The state of the algorithm at this point is
illustrated in Table~\ref{tab:second_sink}. In the next iteration
$\fv{\varsubset_{3,1}}{3}$ remains available as the `best' parent set for node
3 but for node 1 the best parent set now is $\fv{\varsubset_{1,2}}{1}$. In the
second and subsequent iterations the algorithm continues to choose the best
available parent set for some node according to which choice of node
has minimal cost. However in these non-initial iterations cost is
computed as $(\sum_{\varsubset\in\mathrm{ok}(\var)}
\lp{\varsubset}{\var}) -
\lp{\varsubset_{\var,\mathrm{best}}}{\var}$, where
$\fv{\varsubset_{\var,\mathrm{best}}}{\var}$ is the best scoring
remaining choice for \var{} and $\mathrm{ok}(\var)$ is the set
of remaining parent set choices for \var.
 After each such selection, parent set
choices for remaining nodes are updated just like for the first
iteration. Note that the node selected at any iteration will be a sink
node (in the final DAG) for the subset of nodes available at that point.

There is an added complication if some $\fv{\varsubset}{\var}$ are
already fixed to 1 when the sink-finding algorithm is run. This can
happen either due to user constraints or due to branching on
$\fv{\varsubset}{\var}$. Trying to rule out such a variable leads the
algorithm to abort.

Due to its greedy nature the sink finding algorithm is very fast and
so we can afford to run it after solving \emph{every} LP
relaxation. For example, in one of our bigger examples,
\texttt{Diabetes\_100}, the sink-finding
algorithm was called 9425 times taking only 30s in total. Note that
each new batch of cutting planes produces a new LP and thus a new LP
solution. In this way we use the LP to help us move around to search
for high-scoring BNs.

\section{Tightening the LP relaxation}
\label{sec:tighter}

Given a collection of \dimdag{} $\fv{\varsubset}{\var}$ variables,
each feasible DAG corresponds to a (binary) vector in
$\mathbb{R}^{\dimdag}$. Consider now \hull{}, the \emph{convex hull}
of these points and recall \clusterhull{}, the polytope defined in
Section~\ref{sec:branchandcut} containing all points satisfying the
variable bounds, the convexity constraints (\ref{eq:convexity}) and
all cluster constraints (\ref{eq:cluster}). As Jaakkola \emph{et al}
\cite{jaakkola10:_learn_bayes_networ_struc_lp_relax} note, \hull{} is
strictly contained within \clusterhull{}, except in certain special
cases. \gobnilp{} uses \scip{} to add Gomory, strong
CG and zero-half cutting planes in addition to cluster
cuts. This produces a linear relaxation which is tighter than
\clusterhull{} and, as the results presented in
Section~\ref{sec:results} show, typically improves overall performance, although
strong CG cuts are generally not helpful. Denote the
polytope defined by adding these `extra' cuts by \clusterhullscip. Since
the searches for Gomory, strong CG and zero-half cuts
are not complete \clusterhullscip{} is specific to the problem instance and \scip{}
parameter settings.

In many cases it is not possible to separate a fractional
LP solution \lpsol{} even with these extra cuts, so we have
$\lpsol \not\in \hull$ but $\lpsol \in \clusterhullscip$.
 This raises the question of which inequalities are
needed to define \hull{}. We have approached this issue by carrying
out empirical investigations into \hull{} when there are 3 or 4 
nodes.

For 3 nodes $\{1,2,3\}$ there are only 25 DAGs. We eliminated the
three $\fv{\emptyset}{\var}$ variables using the equations
(\ref{eq:convexity}) and encoded each DAG using the remaining
nine $\fv{\varsubset}{\var}$ variables (3 remaining choices of parent
set for each node). The \lrs{} algorithm \cite{lrs} (
\url{http://cgm.cs.mcgill.ca/~avis/C/lrs.html} ) was used to find the
facets of the convex hull of these 25 vertices in
$\mathbb{R}^{9}$.

Denoting this convex hull as $\hull_{3}$, we find that it has 17
facets. These are: 9 lower bounds on the variables, 3 inequalities
corresponding to the original convexity constraints, cluster
constraints for the 4 clusters $\{1,2\}$, $\{1,3\}$, $\{2,3\}$, and
$\{1,2,3\}$ and one additional constraint:
\begin{equation}
  \label{eq:setpacking}
  \fv{\{2,3\}}{1} +   \fv{\{1,3\}}{2} + \fv{\{1,2\}}{3} \leq 1
\end{equation}

Consider the point \lpsolhalf{} in $\mathbb{R}^{9}$ specified by setting
$\fv{\{2,3\}}{1}=\frac{1}{2}$, $\fv{\{1,3\}}{2}=\frac{1}{2}$, $\fv{\{1,2\}}{3}=\frac{1}{2}$
and all other variables to zero. It is not difficult to see that this
point is on the surface of $\clusterhull$ (lying on the
hyperplanes defined by the cluster constraints for $\{1,2\}$,
$\{1,3\}$ and $\{2,3\}$). However it does not satisfy
(\ref{eq:setpacking}) and so is outside of $\hull_{3}$. Note that there
are nine 3-node DAGs where one node has two parents. All of these DAGs
lie on the hyperplane defined by (\ref{eq:setpacking}). It is easy to
show that these 9 DAGs ( = points in $\mathbb{R}^{9}$) are affinely
independent which establishes that (\ref{eq:setpacking}) is indeed a facet.

The point \lpsolhalf{} was also discussed by Jaakkola \emph{et al}
\cite{jaakkola10:_learn_bayes_networ_struc_lp_relax}. They considered
the \emph{acyclic subgraph polytope} \daghull{} which is the convex
hull of DAGs which results from using binary variables to represent
\emph{edges} rather than parent set choices. This polytope has been
extensively studied in discrete mathematics \cite{pdag} and many (but
not all) classes of facets are known for it
\cite{Goemans96thestrongest}. The parent set representation can be
projected onto the edge representation (so that the former is an
\emph{extended formulation} of the latter in the language of
mathematical programming). As Jaakkola \emph{et al} observe the
projection of \lpsolhalf{} is a member of \daghull.

The inequality (\ref{eq:setpacking}) can be generalised to give a class
of valid \emph{set packing} inequalities:
\begin{equation}
  \label{eq:spc}
    \forall \cluster \subseteq V: \sum_{\var \in \cluster}
    \sum_{\varsubset: \cluster \setminus \{\var\} \subseteq \varsubset} \fv{\varsubset}{\var} \leq 1
\end{equation}
We have found that adding all non-trivial
inequalities of this sort for $|C| \leq 4$ speeds up solving
considerably (see Section~\ref{sec:results}). This is because the LP
relaxation is tighter. Since there are not too many such inequalities
they are added directly to the IP rather than being added as cutting planes.
Note that making the connection to characteristic imsets with
(\ref{eq:imset}) implies these set packing constraints.

We have not found all facets of $\hull_{4}$, which is a polytope in
$\mathbb{R}^{28}$ with 543 vertices (for the 543 4-node DAGs). We terminated \lrs{} after a
week's computation, by which time it had found 64 facets. We detected
10 different types of facets which we have labelled 4A to 4J.
We will
provide a full description of these facet classes in a forthcoming
technical report. Here we just give a brief overview.

Consider, as an example, 4B-type facets. They are specified as
follows:
\begin{eqnarray}
\sum_{v_{4}\in\varsubset\wedge\{v_{2},v_{3}\}\cap\varsubset\neq\emptyset}\fv{\varsubset}{v_{1}}
&& \nonumber \\
 +
\sum_{v_{3}\in\varsubset \vee
  \{v_{1},v_{4}\}\subseteq\varsubset}\fv{\varsubset}{v_{2}} && \nonumber \\
+
\sum_{v_{2}\in\varsubset \vee
  \{v_{1},v_{4}\}\subseteq\varsubset}\fv{\varsubset}{v_{3}} && \nonumber \\
+
\sum_{v_{1}\in\varsubset\wedge\{v_{2},v_{3}\}\cap\varsubset\neq\emptyset}\fv{\varsubset}{v_{4}}
  & \leq & 2
  \label{eq:convex4b}
\end{eqnarray}

Consider the point $\lpsolfour \in \mathbb{R}^{28}$ where all
variables take zero value except: $\fv{\{3,4\}}{1}=\frac{1}{2}$,
$\fv{\{1,3\}}{2}=\frac{1}{2}$, $\fv{\{1,4\}}{2}=\frac{1}{2}$,
$\fv{\{2,4\}}{3}=\frac{1}{2}$, $\fv{\{1,2\}}{4}=\frac{1}{2}$. It is
easy to check that $\lpsolfour$ satisfies all cluster constraints and
any constraint of type (\ref{eq:spc}). However, setting
$v_{i}=i$ in (\ref{eq:convex4b}) we have that the left-hand side
is
$2\frac{1}{2}$ and so (\ref{eq:convex4b}) separates
(i.e.\ cuts off) \lpsolfour. It follows that adding 4B-type linear
inequalities results in a strictly tighter linear relaxation.

We have implemented 6 distinct cutting plane algorithms to search for
inequalities of types 4B, 4C, 4E, 4F, 4G and 4H. We call cuts of this
sort \emph{convex4} cuts. Type 4A cuts are
cluster cuts, and cuts of type 4D, 4I and 4J have not appeared useful
in preliminary experiments. We have also experimented with
adding a limited number of convex4 cuts directly to the IP
rather than finding them `on the fly' as cutting planes.

In practice we have found LP solutions which violate these constraints
but which none of our other cutting planes can separate (the example
\lpsolfour{} was extracted from one such LP solution). Cuts of type 4B
appear to be particularly useful. Preliminary experiments indicate
that using convex4 cuts is typically but not always beneficial. We
have yet to do a controlled evaluation, but using convex4 cuts with a
different version of \scip{} (\scip~3.0) and a slightly different
machine from that used to present our main results, we do have some
partial preliminary results. Using convex4 cuts, problem instances
\verb+alarm_3_10000+, \verb+carpo_3_100+, \verb+carpo_3_1000+,
\verb+carpo_3_10000+ and \verb+Diabetes_2_100+ took 54s, 612s, 92s,
660s and 1393s to solve, respectively. All of these times are better
than using our properly evaluated system \newgobnilp{} (see
Section~\ref{sec:results}). The improvement is particular dramatic for
\verb+alarm_3_10000+ which takes 298s using \newgobnilp{} and took
12872s using the system presented by Cussens
\cite{cussens11:_bayes_networ_learn_cuttin_planes}. On the other hand,
problems \verb+hailfinder_3_1+ and \verb+Pigs_2_1000+ took 139s and
2809s respectively which is slower than \newgobnilp.

\section{Set covering and knapsack representations}
\label{sec:setcover}

If the convexity constraints (\ref{eq:convexity}) are all relaxed to set
covering constraints:
$
    \forall \var: \sum_{\varsubset} \fv{\varsubset}{\var} \geq 1
$
then the BN learning problem becomes a pure set covering
problem---albeit one with exponentially many constraints---since all
the cluster constraints (\ref{eq:cluster}) are already set covering
constraints. It is not difficult to show that any optimal solution to
the set covering relaxation of our BN learning problem is also an
optimal solution to the original problem.
This opens up the possibility of applying known
results concerning the set covering polytope to the BN learning
problem. In particular, Balas and Ng
\cite{balas89:_set_cover_polyt} provide conditions for set covering
inequalities to be facets
and also show that for an inequality with integer
coefficients and right-hand side of 1 to be a facet it must be one of the set
covering inequalities defining the IP. This is a useful result for BN
learning. It shows there is no point looking for `extra' set covering
inequalities in the hope they might be facets. Relaxed convexity
constraints and cluster constraints are the only set covering
inequalities that can be facets.

As Balas and Ng \cite{balas89:_set_cover_polyt} note, Chv\'{a}tal's
procedure \cite{chvatal73:_edmon} can be used to generate the convex
hull of integer points satisfying an IP after a finite number of
applications. They provide a specialised version of this procedure to
find a class of valid inequalities of the form $\alpha^{S} x \geq 2$
for the set covering polytope. These inequalities dominate all other
valid inequalities of the form $\beta x \geq 2$. Each such inequality
is defined by taking a set $S$ of set covering
inequalities, and combining them to get a new inequality (details
omitted for space, see \cite{balas89:_set_cover_polyt}).


 We can use the procedure to get new inequalities for the BN learning
 problem by combining cluster
constraints. For example combining the constraints for $C=\{a,b\}, C=\{a,c\}, C=\{a,d\}$ produces:
$\sum_{W: W\cap \{b,c,d\} = \emptyset} 2  \fv{\varsubset}{a}
+  \sum_{W: 0 < |W\cap \{b,c,d\}| < 3}
\fv{\varsubset}{a}
 +  \sum_{a \not\in W} \fv{\varsubset}{b}
 +  \sum_{a \not\in W} \fv{\varsubset}{c}
 +
\sum_{a \not\in W} \fv{\varsubset}{d} \geq 2
$.


Our BN learning problem can also be reformulated as a
 multi-dimensional 0-1 knapsack problem. Due to the
convexity constraints (\ref{eq:cluster}) we can eliminate each
$\fv{\emptyset}{\var}$ variable by replacing it with the linear
expression $1 - \sum_{\varsubset \neq \emptyset}
\fv{\varsubset}{\var}$. Due to pre-pruning, the objective value of
$\fv{\emptyset}{\var}$ will be lower than that for all other
$\fv{\varsubset}{\var}$ variables and so once the
$\fv{\emptyset}{\var}$ variables have been eliminated the remaining
variables will all have positive objective coefficient. Eliminating
$\fv{\emptyset}{\var}$ variables from the cluster constraints produces
the knapsack constraints (\ref{eq:cluster2}) previously mentioned.



\section{Results}
\label{sec:results}

The system \newgobnilp\ described in the previous sections was implemented using C,
with \scip~2.1.1 used as the constraint solver.  The underlying LP
solver was \cplex~12.5.  Both \scip{} and
\cplex{} are available for free under academic licences.  All experiments
were performed using a single core of a 64-bit Linux machine with a 2.80 GHz 4 core
processor and 7.7 GB of RAM.  A time out limit of 2 hours was imposed
across all experiments after which runs were aborted. Our results can
be reproduced by going to \url{http://www.cs.york.ac.uk/aig/sw/gobnilp}.

Experiments were performed on data taken from a variety of Bayesian networks, with different numbers of observations, $N$, and with different limits, $m$, on the maximum number of nodes considered as the parent set of each node.  The problem sets used are shown in Table~\ref{tab:big_table}.  Several of these problem sets were used in~\cite{cussens11:_bayes_networ_learn_cuttin_planes}, with additional larger networks and parent set sizes being added to assess performance on harder problems.  Local BDeu scores were computed for all experiments external to the systems tested and the times taken to compute and filter these are not included in the presented results. Score computation times ranged from 1 second to 5497 seconds in the longest case, \texttt{diabetes} with $N = 10000$.

The primary experiment in this paper is to assess how long
\newgobnilp\ takes to find the BN with the highest score, and rule out the possibility of finding a BN with a higher score for each dataset.  In particular, we compare how the system with all features introduced in previous sections compared to the earlier IP-based BN learning system presented by~\cite{cussens11:_bayes_networ_learn_cuttin_planes} (henceforth referred to as Cussens 2011).

Additionally, we examine the behaviour of the systems in those situations in which they failed to find the highest scoring BN within the 2 hour time limit.  Integer Programming can be used as an any-time learning algorithm, where a current best solution can be taken at any point during the search, though this may not turn out to be the eventual best solution.  Specifically, our aim here is to examine the examples which reached the two hour solving time limit and determine how close to finding a provably best BN they are at that point.  This gives an idea of how good that system would be for use as an any-time algorithm, and acts as an (imperfect) proxy for comparing how much longer the search process would take to reach completion.


The Cussens 2011 system is not publicly available.  However,
\oldgobnilp\ is available and is closely based on Cussens 2011 with
some inefficiencies taken out.  Therefore, comparisons were performed
against \oldgobnilp\ using exactly the same machine and \scip{} and \cplex{}
versions as those used to run \newgobnilp.  These results are shown in
Table~\ref{tab:big_table}.  As the results reported
in~\cite{cussens11:_bayes_networ_learn_cuttin_planes} are performed on
a broadly similar machine to that used for the current experiments,
times taken directly from that paper are also shown in the table for
illustrative purposes.  Results are not shown in the table for some data sets,
as~\cite{cussens11:_bayes_networ_learn_cuttin_planes} did not study and report them.

The results show that in the vast majority of cases,
\newgobnilp\ outperforms \oldgobnilp , often being 2--3 times faster.
\newgobnilp\ is also never slower than Cussens 2011 and for the larger
examples is usually an order of magnitude quicker.  For example, the
\texttt{alarm 3 10000} data set takes over 3.5 hours to solve using
Cussens 2011, but less than 5 minutes using \newgobnilp.  Some of the
difference in run times between \newgobnilp\ and Cussens 2011 may be
due to different machines being used, however the vastly improved
performance on the larger examples undoubtedly reflects an
overwhelming improvement.

In order to discover which aspects of \newgobnilp\ were leading to
this improvement in performance, a second set of experiments was
conducted in which the performance of the full system was compared to
that resulting from removing parts of the system one at a time.  Three
features were identified as being suitable to remove while still
leaving a system that would still result in the best BN
being found, albeit potentially not as efficiently.  These three
features were

\begin{description}
\item[Set Packing Constraints (Section~\ref{sec:tighter}) ] As these
  (\ref{eq:spc}) are logically implied by the basic problem, without them the IP for finding the BN is still correct.
\item[Sink Primal Heuristic (Section~\ref{sec:sinks})] The algorithm for finding feasible solutions through sink finding is not necessary, but may improve the search process through tightening the lower bound on the best BN.
\item[Value Propagator] Explicitly determining which values must be fixed at zero or one at each node of the search tree is not necessary as this information will eventually be discovered through search.  However, by performing this propagation as early as possible, a significant amount of search may be saved.
\end{description}

In addition, three cutting plane algorithms which are built into
\scip\ are used within \newgobnilp.  These three were chosen from
those available in \scip\ based on preliminary experiments to
determine which potential cuts would be added reasonably often and
reasonably quickly.  Each of these was also turned off in turn in
order to assess whether it was positively contributing to the improved
performance.

\begin{sidewaystable*}
\setlength{\tabcolsep}{3pt}
\renewcommand{\arraystretch}{1.3}
\centering
\begin{tabular}{p{1.5cm}p{1cm}p{1cm}p{1.4cm}p{1.4cm}|p{1.4cm}p{1.4cm}p{1.4cm}|p{1.4cm}p{1.4cm}p{1.4cm}|p{1.4cm}p{1.4cm}p{1.4cm}|p{1.4cm}}
\multirow{2}{*}{Network} & \multirow{2}{1.4cm}{\centering $m$} & \multirow{2}{1.4cm}{\centering $p$} & \multirow{2}{1.4cm}{\centering $N$} & \multirow{2}{1.4cm}{\centering Families} &
\multirow{2}{1.4cm}{\centering \newgobnilp} & \multirow{2}{1.4cm}{\centering \oldgobnilp} & \multirow{2}{1.4cm}{\centering Cussens 2011} &
\multicolumn{3}{c|}{Without Solver Feature} & \multicolumn{3}{c|}{No Cuts of Type} &
\multirow{2}{1.4cm}{\centering New VP}\\
&&&&&&&& \centering SPC & \centering SPH & \centering VP & \centering G & \centering SCG & \hspace{0.4cm} ZH \\
\hline
\hline
\multirow{3}{*}{hailfinder} & \centering \multirow{3}{*}{3} & \centering \multirow{3}{*}{56} & \hfill 100 & \hfill 244 & \hfill 1 & \hfill \textit{3} & \hfill 1 & \hfill 1 & \hfill 1 & \hfill 1 & \hfill 1 & \hfill 1 & \hfill 1& \hfill 1\\
&&& \hfill 1000 & \hfill 761 & \hfill 5 & \hfill \textit{14} & \hfill 5 & \hfill \textit{11} & \hfill 5 & \hfill \textbf{4} & \hfill \textbf{4} & \hfill \textbf{4} & \hfill \textbf{4}& \hfill 5\\
&&& \hfill 10000 & \hfill 3708 & \hfill 100 & \hfill \textit{361} & \hfill \textit{169} & \hfill 102 & \hfill \textbf{56} & \hfill 102 & \hfill \textit{558} & \hfill \textbf{75} & \hfill \textbf{83}& \hfill 98\\
\hline
\multirow{3}{*}{hailfinder} & \centering \multirow{3}{*}{4} & \centering \multirow{3}{*}{56} & \hfill  100 & \hfill 4106 &\hfill 18 & \hfill \textit{270} &&\hfill \textit{34} &\hfill 18 &\hfill \textit{34} &\hfill 18 &\hfill \textbf{13} &\hfill \textbf{13}& \hfill 19\\
&&& \hfill 1000 & \hfill 767 &\hfill 4 & \hfill \textit{14} &&\hfill \textit{10} &\hfill 4 &\hfill 4 &\hfill \textit{5} &\hfill \textit{6} &\hfill 4& \hfill 4\\
&&& \hfill 10000 & \hfill 4330 &\hfill 68 & \hfill \textit{934} &&\hfill 62 &\hfill \textit{128} &\hfill \textit{587} &\hfill \textit{216} &\hfill 71 &\hfill 71& \hfill 70\\
\hline
\multirow{3}{*}{alarm} & \centering \multirow{3}{*}{3} & \centering \multirow{3}{*}{37} & \hfill 100 & \hfill 907 & \hfill 2 & \hfill \textit{6} & \hfill \textit{4} & \hfill \textit{3} & \hfill 2 & \hfill \textbf{1} & \hfill 2 & \hfill 2 & \hfill 2& \hfill 2\\
&&& \hfill 1000 & \hfill 1928 & \hfill 5 & \hfill \textit{14} & \hfill \textit{15} & \hfill \textit{12} & \hfill \textbf{4} & \hfill 5 & \hfill \textit{7} & \hfill \textbf{4} & \hfill 5& \hfill \textbf{4}\\
&&& \hfill 10000 & \hfill 6473 & \hfill 289 & \hfill \textit{792} & \hfill \textit{12872} & \hfill \textit{479} & \hfill \textit{394} & \hfill \textit{397} & \hfill \textit{739} & \hfill \textit{1049} & \hfill \textit{710}& \hfill 298\\
\hline
\multirow{3}{*}{alarm} & \centering \multirow{3}{*}{4} & \centering \multirow{3}{*}{37} & \hfill  100 & \hfill 1293 &\hfill 2 & \hfill \textit{7} &&\hfill \textit{9} &\hfill 2 &\hfill \textit{7} &\hfill \textit{3} &\hfill 2 &\hfill 2& \hfill 2\\
&&& \hfill  1000 & \hfill 2097 &\hfill 7 & \hfill \textit{15} &&\hfill \textit{21} &\hfill \textbf{6} &\hfill \textit{8} &\hfill \textbf{6} &\hfill \textbf{3} &\hfill \textbf{6}& \hfill 7\\
&&& \hfill  10000 & \hfill 8445 &\hfill 398 & \hfill \textit{839} &&\hfill \textit{1253} &\hfill \textit{806} &\hfill \textit{1421} &\hfill \textit{633} &\hfill \textit{1567} &\hfill \textit{1052}& \hfill 398\\
\hline
\multirow{3}{*}{carpo} & \centering \multirow{3}{*}{3} & \centering \multirow{3}{*}{60} & \hfill 100 & \hfill 5068 & \hfill 756 & \hfill \textit{887} & \hfill \textit{15176} & \hfill 742 & \hfill \textbf{628} & \hfill 716 & \hfill 690 & \hfill \textbf{642} & \hfill 740& \hfill \textbf{651}\\
&&& \hfill 1000 & \hfill 3827 & \hfill 106 & \hfill \textit{171} & \hfill \textit{593} & \hfill \textit{143} & \hfill \textit{134} & \hfill 115 & \hfill \textit{117} & \hfill 104 & \hfill 110& \hfill 99\\
&&& \hfill 10000 & \hfill 16391 & \hfill 1311 & \hfill \textbf{566} & \hfill \textit{42275} & \hfill \textit{2158} & \hfill \textit{1574} & \hfill \textbf{1071} & \hfill \textit{4350} & \hfill \textbf{1032} & \hfill \textit{2057}& \hfill 1286\\
\hline
\multirow{3}{*}{carpo} & \centering \multirow{3}{*}{4} & \centering \multirow{3}{*}{60} & \hfill  100 & \hfill 13185 &\hfill [0\%] & \hfill [\textit{1}\%] &&\hfill 6649 &\hfill [0\%] &\hfill [0\%] &\hfill [0\%] &\hfill 7014 &\hfill [0\%]& \hfill  [0\%]\\
&&& \hfill  1000 & \hfill 4722 &\hfill 151 & \hfill \textit{406} &&\hfill \textit{252} &\hfill \textit{168} &\hfill \textit{188} &\hfill \textit{240} &\hfill \textit{208} &\hfill 140& \hfill 149\\
&&& \hfill  10000 & \hfill 34540 &\hfill [0\%] & \hfill 4065 &&\hfill [0\%] &\hfill [0\%] &\hfill [0\%] &\hfill [0\%] &\hfill [0\%] &\hfill [0\%]& \hfill  [0\%]\\
\hline
\multirow{3}{*}{diabetes} & \centering \multirow{3}{*}{2} & \centering \multirow{3}{*}{413} & \hfill 100 & \hfill 4441 & \hfill 2982 & \hfill [\textit{31}\%] && \hfill [\textit{39}\%] & \hfill 3082 & \hfill 3040 & \hfill 3036 & \hfill \textbf{1506} & \hfill 3212& \hfill 2745\\
&&& \hfill 1000 & \hfill 21493 & \hfill [17\%] & \hfill [\textit{23}\%] &&  \hfill [\textit{168}\%] & \hfill [\textit{199}\%] & \hfill [16\%] & \hfill [17\%] & \hfill [\textbf{15}\%] & \hfill [17\%]& \hfill  [18\%]\\
&&& \hfill 10000 & \hfill 262129 & \hfill [44\%] & \hfill [\textbf{17}\%] && \hfill [\textit{378}\%] & \hfill [\textit{380}\%] & \hfill [44\%] & \hfill [44\%] & \hfill [44\%] & \hfill [44\%]& \hfill  [44\%]\\
\hline
\multirow{3}{*}{pigs} & \centering \multirow{3}{*}{2} & \centering \multirow{3}{*}{441} & \hfill 100 & \hfill 2692 & \hfill 89 & \hfill [\textit{0}\%] && \hfill \textit{5103} & \hfill 87 & \hfill \textbf{32} & \hfill 88 & \hfill 85 & \hfill 89& \hfill \textbf{32}\\
&&& \hfill 1000 & \hfill 15847 & \hfill 1818 & \hfill [\textit{7}\%] && \hfill [\textit{8}\%] & \hfill 1788 & \hfill 1715 & \hfill 1714 & \hfill 1802 & \hfill 1822& \hfill 1657\\
&&& \hfill 10000 & \hfill 304219 & \hfill [3\%] & \hfill [\textit{9}\%] && \hfill [\textit{13}\%] & \hfill [\textit{42}\%] & \hfill [3\%] & \hfill [3\%] & \hfill [3\%] & \hfill [3\%]& \hfill  [3\%]\\
\end{tabular}
\caption{\label{tab:big_table} Comparison of \newgobnilp\ with older systems and impact of various features.
         $p$ is the number of variables in the data set.
         $m$ is the limit on the number of parents of each variable.
         $N$ is the number of observations in the data set.
         Families is the number of family variables in the data set after pruning.
         All times are given in seconds (rounded).
         ``[---]'' indicates that the solution had not been found
         after 2 hours --- the value given is the gap, rounded to the
         nearest percent, between the score of the best found BN and the upper bound on the score of the best potential BN, as a percentage of the score of the best found BN.
         Entries in italics are at least 10\% worse than \newgobnilp, while those in bold are at least 10\% better.
         Key: SPC -- Set Packing Constraints, SPH -- Sink Primal Heuristic, VP -- Value Propagator, G -- Gomory cuts, SCG -- Strong CG cuts, ZH -- Zero-half cuts.}
\end{sidewaystable*}

Six modified versions of the \newgobnilp\ resulted from this; three which each had one feature turned off and three which each had one type of cutting plane turned off.  Each of the data sets was run on each of these systems and the time taken to find the optimal solution recorded.  The results of these experiments are shown in Table~\ref{tab:big_table}.

The results show the biggest change in solution time occurs when the
set packing constraints are removed.  In nearly every case, this leads
to an increase in solution time.  In fact the situation can be even more extreme than the
table suggests; for the \texttt{pigs 1000} data set, the system
without the set packing constraints was allowed to continue running
beyond the time out limit and had still not finished after more than
30 hours, when the full system finished in about 30 minutes.

Furthermore, in cases in which the two hour time limit was reached, the gap between the upper and lower bounds at that point was much larger in the version without set packing constraints than that for the full system.  Closer examination revealed that this was because the score of the best BN found so far was the
same as in the full system, but significantly less progress had been
made in reducing the upper (dual) bound.

It is not immediately clear from this table if using the sinks heuristic aids the system or not.  There are a number of problems for which it decreases solution time and a number for which it increases it.  For the most difficult problems, it appears to have little impact on solving time.  As with the version without set packing constraints, the system without the primal heuristic also resulted in much larger gaps between the bounds in cases where it reached the time out limit. For the version without the primal heuristic, the difference
is due to the best BN found so far being significantly worse after the
time had elapsed, while the upper bound on the best possible BN that
could be found was virtually identical to the full system.  This
latter point suggests that, while the sinks heuristic was of
questionable value when solving problems to completion, on larger
problems for which the algorithm may run out of time or resources
before successfully finding the provably best BN, the heuristic plays
an import role in ensuring that the best BN found so far is of high
score.

The evidence for the effectiveness of the propagation is also mixed.  In some cases, the system performs faster without the propagation, though study of the log files reveals this is almost exclusively down to the time spent directly carrying out propagation.  Having noted this, a new faster propagator was created and used to replace the existing one.  The result of running the full system with this faster propagator was to achieve a minor improvement over both the full system and the system without a propagator, as shown in the final column of Table~\ref{tab:big_table}.

The usefulness of each of the cutting plane algorithms is much clearer.  Without Gomory cuts, the system is often slower and frequently by a large amount.  On the other hand, removing Strong CG cuts usually improves the system's performance with one notable exception.  Though checking log files, it can be confirmed that little time is spent generating Strong CG cuts and hence the improvement without them is due to a better search strategy rather than just a saving on the time spent adding these cuts.  Zero-half cuts appear to have less effect than the other two studied but reduce solving time noticeably on two fairly large problems.

\section{Conclusions and future work}
\label{sec:conclusions}

Our principal conclusion is that IP is an attractive framework for
exact BN learning from complete discrete data. However, as our
comparative experiments demonstrate, some care (and empirical
investigation) is required to properly exploit its potential. We have
shown considerable advances over the results presented by Cussens
\cite{cussens11:_bayes_networ_learn_cuttin_planes} who himself
presented faster solving times on four problems than
\cite{jaakkola10:_learn_bayes_networ_struc_lp_relax}. However, it
would clearly be desirable to compare \newgobnilp{} against further
exact BN learning systems, not necessarily IP-based. We intend to
compare against the \urlearning{} system \cite{Yuan12improved} in the
immediate future.

In this paper we have focused on efficiently finding BNs with maximal
score subject to constraints on parent set size. This raises the
question of whether it is worth the effort to do this if one's
ultimate goal is to return a DAG with high structural accuracy. In the
context of `pedigree reconstruction', work by Cussens \emph{et al}
\cite{cussens13:_maxim_likel_pedig_recon_integ_linear_progr} answers
this question in the positive. In that paper an exact learning
approach led to high structural accuracy. However, in a pedigree ( a
`family tree') no node can have more than two parents. In other
applications where the `true' structure may have nodes with many
parents our current restriction on parent set size may lead to poor
structural accuracy.  This is a clear limitation which we intend to
address by the IP technique of \emph{delayed column generation} where
IP variables (i.e.\ parent sets) are created during solving
\cite{cussens12:_colum_bn}.

In practical applications one often has prior knowledge concerning the
(likely) structure of the `true' BN. Because our \newgobnilp{} system
is an example of \emph{declarative machine learning} it is very easy
to allow the user to declare constraints on BN structure when these
can be encoded as linear constraints. Although we have not exploited it
in the experiments reported here, \newgobnilp{} allows the user to
declare the absence/presence of (i) particular directed edges, (ii)
particular adjacencies and (iii) particular immoralities. In addition
upper and lower bounds on the number of edges and founder nodes are
possible. It is also possible to rule out specific BNs with a
linear constraint. This allows \newgobnilp{} to not only find the
optimal BN but also the top $k$ BNs in decreasing order of score.


\section*{Acknowledgements}
\label{sec:ack}

Thanks  to Milan Studen\'{y}, Raymond Hemmecke
and David Haws for useful discussions concerning the imset representation.
We would like to thank three anonymous referees for their valuable
comments. This work has been supported by the UK Medical Research
Council (Project Grant G1002312). This work benefited from a visit by
the second author to the Helsinki Institute for Information Technology
supported by the Finnish Centre of Excellence in Computational
Inference Research (COIN).

\bibliographystyle{plain}
\bibliography{submit}

\end{document}